\documentclass[11pt,a4paper]{article}
\usepackage[hyperref]{emnlp2018}
\usepackage{times}
\usepackage{latexsym}
\usepackage[colorinlistoftodos]{todonotes}

\usepackage{url}
\usepackage{multirow}
\usepackage{mathtools}
\usepackage{amsmath,graphicx}
\usepackage[small]{caption}
\usepackage{xcolor}
\usepackage{xspace}
\usepackage{paralist}
\usepackage{url}
\usepackage{enumitem}
\usepackage{multicol}
\usepackage[colorinlistoftodos]{todonotes}
\usepackage{subcaption}
\usepackage[export]{adjustbox}
\setlist[itemize]{leftmargin=*}
\usepackage{footnote}
\makesavenoteenv{tabular}
\makesavenoteenv{table}

\usepackage[symbol]{footmisc}

\renewcommand{\paragraph}[1]{\medskip\noindent\textbf{#1}~}

\newcommand{\namedref}[2]{\hyperref[#2]{#1~\ref*{#2}}}

\newcommand{\sectionref}[1]{\namedref{Section}{sec:#1}}
\newcommand{\tableref}[1]{\namedref{Table}{tab:#1}}
\newcommand{\figureref}[1]{\namedref{Figure}{fig:#1}}

\newcommand{\DLM}{D-RNNLM\xspace}
\newcommand{\out}{\ensuremath{O}\xspace}
\newcommand{\stt}{\ensuremath{H}\xspace}
\newcommand{\dummy}{\ensuremath{\#}\xspace}

\aclfinalcopy 

\title{Code-switched Language Models \\ Using Dual RNNs and Same-Source Pretraining} 

\author{Saurabh Garg\thanks{\, Joint first authors} \And Tanmay Parekh\footnotemark[1]\\\\ Indian Institute of Technology, Bombay \\ \texttt{\{saurabhgarg,tanmayb,pjyothi\}@cse.iitb.ac.in} \And Preethi Jyothi}

\date{}

\begin{document}
\maketitle
\begin{abstract}
  This work focuses on building language models (LMs) for code-switched text. We propose two techniques that significantly improve these LMs: 1) A novel recurrent neural network unit with dual components that focus on each language in the code-switched text separately 2) Pretraining the LM using synthetic text from a generative model estimated using the training data. We demonstrate the effectiveness of our proposed techniques by reporting perplexities on a Mandarin-English task and derive significant reductions in perplexity. 
\end{abstract}
\section{Introduction}

Code-switching is a widespread linguistic phenomenon among multilingual speakers that involves switching between two or more languages in the course of a single conversation or within a single sentence~\cite{auer2013code}. Building speech and language technologies to handle code-switching has become a fairly active area of research and presents a number of interesting technical challenges~\cite{ccetinoglu2016challenges}. 
Language models for code-switched text is an important problem with implications to downstream applications such as speech recognition and machine translation of code-switched data.  A natural choice for building such language models would be to use recurrent neural networks (RNNs)~\cite{mikolov2010recurrent}, which yield state-of-the-art language models in the case of monolingual text. In this work, we explore mechanisms that can significantly improve upon such a baseline when applied to code-switched text. Specifically, we develop two such mechanisms:
\begin{itemize}
\item We alter the structure of an RNN unit to include separate components that focus on each language in code-switched text separately while coordinating with each other to retain contextual information across code-switch boundaries. Our new model is called a {\em Dual RNN Language Model} (\DLM), described in \sectionref{dlm}.
\item We propose using {\em same-source pretraining} -- i.e., pretraining the model using data sampled from a generative model which is itself trained on the given training data -- before training the model on the same training data (see \sectionref{textgen}). We find this to be a surprisingly effective strategy.
\end{itemize}

We study the improvements due to these techniques under various settings (e.g., with and without access to monolingual text in the candidate languages for pretraining). We use perplexity as a proxy to measure the quality of the language model, evaluated on code-switched text in English and Mandarin from the SEAME corpus. Both the proposed techniques are shown to yield {\em significant perplexity improvements (up to 13\% relative)} over different baseline RNNLM models (trained with a number of additional resources). We also explore how to combine the two techniques effectively.


\paragraph{Related Work:} \newcite{adel2013recurrent} was one of the first works to explore the use of RNNLMs for code-switched text. Many subsequent works explored the use of external sources to enhance code-switched LMs, including the use of part-of-speech (POS) tags, syntactic and semantic features~\cite{yeh2010integrated,adel2014combining, adel2015syntactic} and the use of machine translation systems to generate synthetic text~\cite{vu2012first}. Prior work has also explored the use of interpolated LMs trained separately on monolingual texts~\cite{bhuvanagiri2010approach,imseng2011language,li2011asymmetric, baheti2017curricula}. 
Linguistic constraints governing code-switching have also been used as explicit priors to model when people switch from one language to another. Following this line of enquiry,~\cite{zhang2008grammar} used grammar rules to model code-switching;~\cite{li2013improved,li2014functional} incorporated syntactic constraints with the help of a code-switch boundary prediction model;~\cite{pratapa2018language} used a linguistically motivated theory to create grammatically consistent synthetic code-mixed text.

\begin{figure}[t!]
\includegraphics[width=1.05\linewidth]{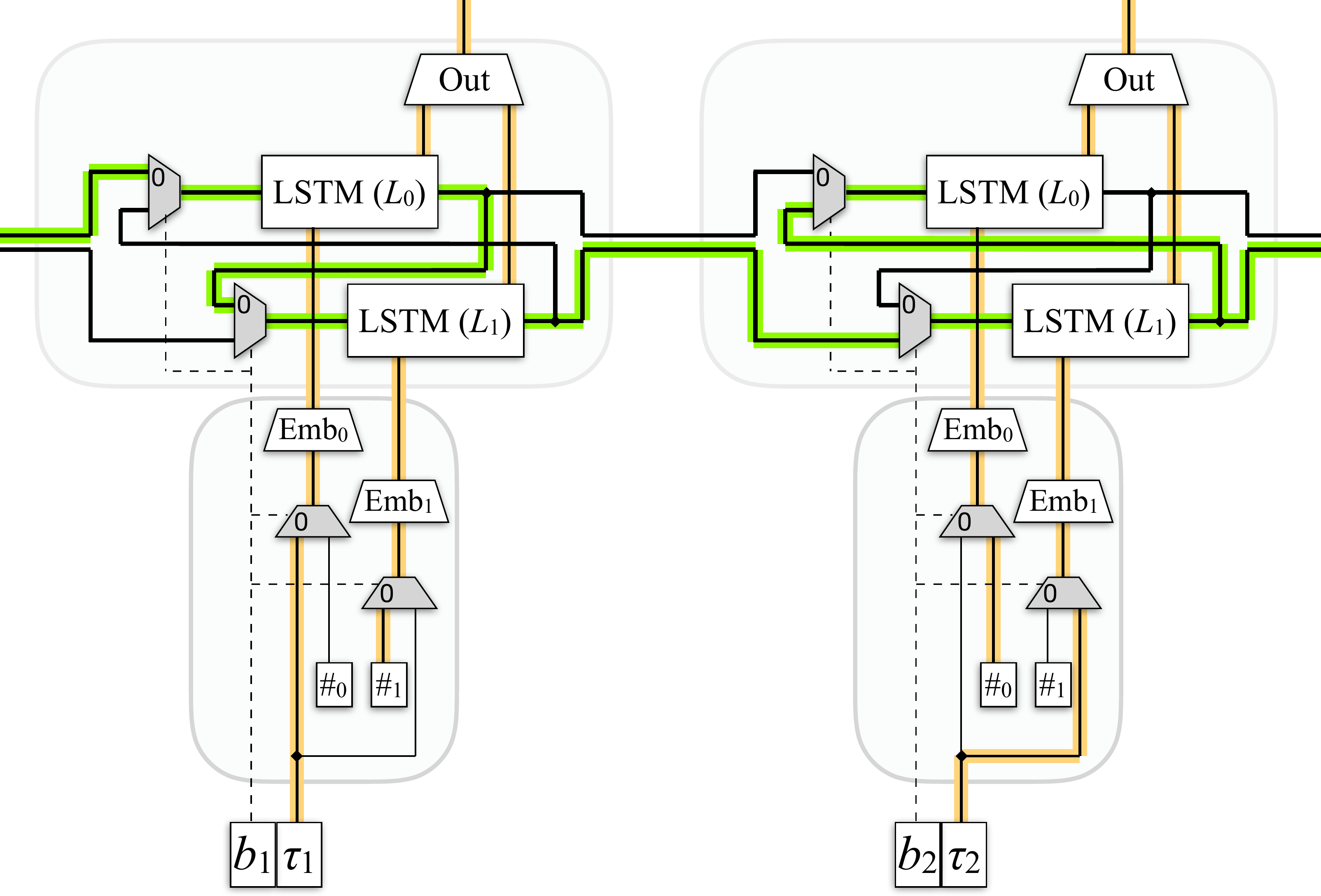}
\caption{Illustration of the dual RNNLM (see the text for a detailed description). The highlighted left-to-right path (in green) indicates the flow of state information, when $b_1=0$ and $b_2=1$ (corresponding to token $\tau_1$ belonging to language $L_0$ and $\tau_2$ belonging to $L_1$). The highlighted bottom-to-top path (in orange) indicates the inputs and outputs.}
\label{fig:DLM}
\end{figure}
\section{Dual RNN Language Models}
\label{sec:dlm}

Towards improving the modeling of code-switched text, we introduce Dual RNN Language Models (\DLM{s}). The philosophy behind \DLM{s} is that two different sets of neurons will be trained to (primarily) handle the two languages. (In prior work~\cite{garg2018}, we applied similar ideas to build dual N-gram based language models for code-switched text.)

As shown in \figureref{DLM}, the \DLM consists of a ``Dual LSTM cell'' and an input encoding layer.  The Dual LSTM cell, as the name indicates, has two long short-term memory (LSTM) cells within it. The two LSTM cells are designated to accept input
tokens from the two languages $L_0$ and $L_1$ respectively, and produce an (unnormalized) output distribution over the tokens in the same language. When a Dual LSTM cell is invoked with an input token $\tau$, the two cells
will be invoked sequentially. The first (upstream) LSTM cell corresponds to
the language that $\tau$ belongs to, and gets $\tau$ as its input. It passes on the resulting state to the downstream LSTM cell (which takes a dummy token as input). The unnormalized outputs from the two cells are combined and passed through a soft-max operation to obtain a distribution over the
union of the tokens in the two languages. \figureref{DLM} shows a circuit representation of this configuration, using multiplexers (shaded units) controlled by a selection bit $b_i$ such that the $i^\text{th}$ token $\tau_i$ belongs to $L_{b_i}$.

The input encoding layer also uses multiplexers to direct the input token to the upstream LSTM cell. Two dummy tokens $\#_0$ and $\#_1$ are added to $L_0$ and $L_1$ respectively, to use as inputs to the downstream LSTM cell. The input tokens are encoded using an embedding layer of the network (one
for each language), which is trained along with the rest of the network to minimize a cross-entropy loss function.

The state-update and output functions of the Dual LSTM cell can be formally described as follows. It takes as input $(b,\tau)$ where $b$ is a bit and $\tau$ is an input token, as well as a state vector of the form $(h_0,h_1)$ corresponding to the state vectors produced by its two constituent LSTMs. Below we denote 
the state-update and output functions of these two LSTMs as $\stt_b(\tau,h)$ and $\out_b(\tau,h)$ (for $b=0,1$):
\begin{gather*}
\stt((b,\tau),(h_0,h_1)) = (h'_0,h'_1) \qquad\text{where} \\
(h'_0,h'_1) = 
\begin{cases}
(\stt_0(\tau,h_0),\stt_1(\dummy_1,h'_0)) & \text{if } b=0\\
(\stt_0(\dummy_0,h'_1),\stt_1(\tau,h_1)) & \text{if } b=1
\end{cases}\\
\out((b,\tau),(h_0,h_1)) = \mathrm{softmax}(o_0,o_1) \qquad\text{where} \\
(o_0,o_1) = \begin{cases}
(\out_0(\tau,h_0),\out_1(\dummy_1,h'_0)) & \text{if } b=0\\
(\out_0(\dummy_0,h'_1),\out_1(\tau,h_1)) & \text{if } b=1.
\end{cases}
\end{gather*}
Note that above, the inputs to the downstream LSTM functions $\stt_{1-b}$ and $\out_{1-b}$ are expressed in terms of $h'_b$ which is produced by the upstream LSTM.

\section{Same-Source Pretraining}
\label{sec:textgen}

Building robust LMs for code-switched text is challenging due to the lack of availability of large amounts of training data. One solution is to artificially generate code-switched to augment the training data. We propose a variant of this approach -- called same-source pretraining -- in which the actual training data itself is used to train a generative model, and the data sampled from this model is used to pretrain the language model.

Same-source pretraining can leverage powerful training techniques for generative models to train a language model. We note that the generative models by themselves are typically trained to minimize a different objective function (e.g., a discrimination loss) and need not perform well as language models.%
\footnote{In our experiments, we found the preplexity measures for the generative models to be an {\em order of magnitude} larger than that of the LMs we construct.}

Our default choice of generative model will be an RNN (but see the end of this paragraph). To complete the specification of same-source pretraining, we need to specify how it is trained from the given data. Neural language models trained using the maximum likelihood training paradigm tend to suffer from the {\em exposure bias} problem during inference when the model generates a text sequence by conditioning on previous tokens that may never have appeared during training. Scheduled sampling~\cite{bengio2015scheduled} can help bridge this gap between the training and inference stages by using model predictions to synthesize prefixes of text that are used during training, rather than using the actual text tokens. A more promising alternative to generate text sequences was recently proposed by~\newcite{yu2017seqgan} where sequence generation is modeled in a generative adversarial network (GAN) based framework. This model -- referred to as ``SeqGAN'' -- consists of a generator RNN and a discriminator network trained as a binary classifier to distinguish between real and generated sequences. The main innovation of SeqGAN is to train the generative model using policy gradients (inspired by reinforcement learning) and use the discriminator to determine the reward function. We  experimented with using both na\"ive and scheduled sampling based training; using SeqGAN was a consistently better choice (5 points or less in terms of test perplexities) compared to these two sampling methods. As such, we use SeqGAN as our training method for the generator. We also experiment with replacing the RNN with a Dual RNN as the generator in the SeqGAN training and observe small but consistent reductions in perplexity. 
\begin{table}[t!]
\resizebox{\columnwidth}{!}{
\centering
 \begin{tabular}[b]{ | c | c | c | c | }
 \hline
 {} & Train & Dev & Test \\
 \hline\hline
 \# Utterances & $74,927$ & $9,301$ & $9,552$ \\
 \hline
 \# Tokens & $977,751$ & $131,230$ & $114,546$ \\
 \hline
  \# English Tokens & $316,726$ & $30,154$ & $50,537$ \\
 \hline
  \# Mandarin Tokens & $661,025$ & $101,076$ & $64,009$ \\
 \hline
\end{tabular}
}
\caption{Statistics of data splits derived from SEAME.\vspace{-1em}\label{tab:data}}
\end{table}
\begin{table*}[t!]
\small
\centering
 \begin{tabular}{ | c | c | c | c | c | c | c | c | c |}
 \hline
 \multirow{3}{*}{} & \multicolumn{4}{|c|}{w/o syntactic features} & \multicolumn{4}{|c|}{with syntactic features} \\
 \cline{2-9}
 & \multicolumn{2}{|c|}{w/o mono. data} & \multicolumn{2}{|c|}{with mono. data} & \multicolumn{2}{|c|}{w/o mono. data} & \multicolumn{2}{|c|}{with mono. data} \\
  \cline{2-9}
 & Dev & Test & Dev & Test & Dev & Test & Dev & Test \\ 
 \hline
Baseline & 89.60 & 74.87 & 74.06 & 61.66 & 81.87 & 68.23 & 71.04 & 59.00 \\ 
\hline
\DLM & 88.68 & 72.29 & 72.41 & 60.73 & 81.01 & 66.26 & 70.83 & 59.04 \\
\hline
With RNNLM SeqGAN & 79.16 & 65.96 & 72.51 & 60.56 & 77.30 & 63.75 & 68.43 & 55.71 \\
\hline
With \DLM SeqGAN & \textbf{78.63} & \textbf{65.41} & \textbf{72.33} & \textbf{60.30} & \textbf{77.19} & \textbf{63.63} & \textbf{67.79} & \textbf{55.60} \\
\hline
\end{tabular}
\captionof{table}{Development set and test set perplexities using RNNLMs and D-RNNLMs with various pretraining strategies.\vspace{-1em}} \label{tab:results}
\end{table*}
\section{Experiments and Results}
\vspace{-0.5em}
\paragraph{Dataset Preparation:} For our experiments, we use code-switched text from the SEAME corpus~\cite{lyu2010analysis} which contains conversational speech in Mandarin and English. Since there is no standardized task based on this corpus, we construct our own training, development and test sets using a random 80-10-10 split. \tableref{data} shows more details about our data sets. (Speakers were kept disjoint across these datasets.) 

\paragraph{Evaluation Metric:} We use token-level perplexity as the evaluation metric where tokens are words in English and characters in Mandarin. The SEAME corpus provides word boundaries for Mandarin text. However, we used Mandarin characters as individual tokens since a large proportion of Mandarin words appeared very sparsely in the data. Using Mandarin characters as tokens helped alleviate this issue of data sparsity; also, applications using Mandarin text are typically evaluated at the character level and do not rely on having word boundary markers~\cite{vu2012first}.

\paragraph{Outline of Experiments:} Section~\ref{sec:expts1} will explore the benefits of both our proposed techniques -- (1) using D-RNNLMs and (2) using text generated from SeqGAN for pretraining -- in isolation and in combination. Section~\ref{sec:expts2} will introduce two additional resources (1) monolingual text for pretraining and (2) a set of syntactic features used as additional input to the RNNLMs that further improve baseline perplexities. We show that our proposed techniques continue to outperform the baselines albeit with a smaller margin. All these perplexity results have been summarized in~\tableref{results}.

\subsection{Improvements Over the Baseline}
\label{sec:expts1}

This section focuses only on the numbers listed in the first two columns of~\tableref{results}.  The Baseline model is a $1$-layer LSTM LM with $512$ hidden nodes, input and output embedding dimensionality of 512, trained using SGD with an initial learning rate of $1.0$ (decayed exponentially after $80$ epochs at a rate of $0.98$ till 100 epochs) The development and test set perplexities using the baseline are $89.60$ and $74.87$, respectively.%

%

The D-RNNLM is a $1$-layer language model with each LSTM unit having $512$ hidden nodes. The training paradigm is similar to the above-mentioned setting for the baseline model.%
\footnote{\DLM{s} have a few additional parameters. However, increasing the capacity of an RNNLM to exactly match this number makes its test perplexity worse; RNNLM with 720 hidden units gives a development set perplexity of 91.44 and 1024 hidden units makes it 91.46.}
%
We see consistent improvements in test perplexity when comparing a D-RNNLM with an RNNLM (i.e. 74.87 drops to 72.29).\footnote{Since \DLM{s} use language ID information, we also trained a baseline RNNLM with language ID features; this did not help reduce the baseline test perplexities. In future work, we will explore alternate LSTM-based models that incorporate language ID information~\cite{chandu2018language}}


Next, we use text generated from a SeqGAN model to pretrain the RNNLM.%
\footnote{To implement SeqGAN, we use code from \url{https://github.com/LantaoYu/SeqGAN}.}
We use our best trained RNNLM baseline as the generator within SeqGAN. We sample 157,440  sentences (with a fixed sentence length of 20) from the SeqGAN model; this is thrice the amount of code-switched training data. We first pretrain the baseline RNNLM with this sampled text, before training it again on the code-switched text. This gives significant reductions in test perplexity, bringing it down to 65.96 (from 74.87).


Finally, we combine both our proposed techniques by replacing the generator with our best-trained \DLM within SeqGAN. Although there are other ways of combining both our proposed techniques, e.g. pretraining a \DLM using data sampled from an RNNLM SeqGAN, we found this method of combination to be most effective. We see modest but consistent improvements with \DLM SeqGAN over RNNLM SeqGAN in~\tableref{results}, further   validating the utility of \DLM{s}.  

\subsection{Using Additional Resources}
\label{sec:expts2}

We employed two additional resources to further improve our baseline models. First, we used monolingual text in the candidate languages to pretrain the RNNLM and \DLM models. We used transcripts from the Switchboard corpus%
\footnote{http://www.openslr.org/5/}
for English; AIShell%
\footnote{http://www.openslr.org/33/} %
and THCHS30%
\footnote{http://www.openslr.org/18/} %
corpora for Mandarin monolingual text data. This resulted in a total of $\approx$3.1 million English tokens and $\approx$2.5 million Mandarin tokens. Second, we used an additional set of input features to the RNNLMs and \DLM{s} that were found to be useful for code-switching in prior work~\cite{adel2014combining}. The feature set included part-of-speech (POS) tag features and Brown word clusters~\cite{brown1992class}, along with a language ID feature. We extracted POS tags using the Stanford POS-tagger%
\footnote{\url{https://nlp.stanford.edu/software/tagger.shtml}}
and we clustered the words into 70 classes using the unsupervised clustering algorithm by~\newcite{brown1992class} to get Brown cluster features. 

The last six columns in~\tableref{results} show the utility of using either one of these resources or both of them together (shown in the last two columns). The perplexity reductions are largest (compared to the numbers in the first two columns) when combining both these resources together. Interestingly, all the trends we observed in~\sectionref{expts1} still hold. \DLM{s} still consistently perform better than their RNNLM counterparts and we obtain the best overall results using \DLM SeqGAN. 

\section{Discussion and Analysis}
\label{sec:disc}

\begin{table}[ht!]
\resizebox{\columnwidth}{!}{
\centering
 \begin{tabular}[b]{ | c | c | c | c | c | }
 \hline
  & Eng-Eng & Eng-Man & Man-Eng & Man-Man \\
 \hline\hline
  RNNLM & $133.18$ & $157.18$ & $2617.28$ & $34.98$\\
  D-RNNLM & $140.37$ & $151.38$ & $2452.16$ & $32.89$\\
 \hline
  Mono RNNLM & $101.61$ & $181.28$ & $2510.48$ & $30.00$\\
 Mono D-RNNLM & $101.66$ & $156.44$ & $2442.81$ & $29.64$\\
 \hline
  RNNLM SeqGAN & $120.28$ & $154.44$ & $2739.85$ & $30.40$ \\
 D-RNNLM SeqGAN & $120.26$ & $149.68$ & $2450.85$ & $30.60$\\
 \hline
\end{tabular}
}
\captionof{table}{Decomposed perplexities on the development set on all four types of tokens from various models.\vspace{-0.5em} \label{tab:perpanalysis}}
\end{table}
\tableref{perpanalysis} shows how the perplexities on the development set from six of our prominent models decompose into the perplexities contributed by English tokens preceded by English tokens (Eng-Eng), Eng-Man, Man-Eng and Man-Man tokens. This analysis reveals a number of interesting observations. 1) The \DLM mainly improves over the baseline on the ``switching tokens'', Eng-Man and Man-Eng. 2) The RNNLM with monolingual data improves most over the baseline on ``the monolingual tokens'', Eng-Eng and Man-Man, but suffers on the Eng-Man tokens. The \DLM with monolingual data does as well as the baseline on the Eng-Man tokens and performs better than ``Mono RNNLM'' on all other tokens. 3) RNNLM SeqGAN suffers on the Man-Eng tokens, but helps on the rest; in contrast, \DLM SeqGAN helps on {\em all tokens} when compared with the baseline.

\begin{table}[ht!]
\small
\resizebox{\columnwidth}{!}{
\centering
 \begin{tabular}[b]{ | c | c | c | }
 \hline
  & SeqGAN-RNNLM & SeqGAN-DLM \\
 \hline\hline
 Bigram &  $25.57$ & $31.33$\\
 \hline
 Trigram & $75.88$ & $83.86$\\
 \hline
 Quadgram & $137.98$ & $145.71$\\
 \hline
\end{tabular}
}
\captionof{table}{Percentage of new n-grams generated.\vspace{-1em}\label{tab:unique}}
\end{table}
As an additional measure of the quality of text generated by RNNLM SeqGAN and \DLM SeqGAN, in \tableref{unique}, we measure the diversity in the generated text by looking at the increase in the number of unique n-grams with respect to the SEAME training text. \DLM SeqGAN is clearly better at generating text with larger diversity, which could be positively correlated with the perplexity improvements shown in~\tableref{results}.

While we do not claim same-source pretraining may be an effective strategy in general, we show it is useful in low training-data scenarios. Even with only $\frac1{16}$th of the original SEAME training data used for same-source pretraining, development and test perplexities are reduced to 84.45 and 70.59, respectively (compared to 79.16 and 65.96 using the entire training data). 

\section{Conclusion}
\DLM{s} and same-source pretraining provide significant perplexity reductions for code-switched LMs. These techniques may be of more general interest. Leveraging generative models to train LMs is potentially applicable beyond code-switching; \DLM{s} could be generalized beyond LMs, e.g. speaker diarization. We leave these for future work to explore.

\section{Acknowledgments}
The authors thank the anonymous reviewers for their helpful comments and suggestions. The last author gratefully acknowledges financial support from Microsoft Research India (MSRI) for this project.

\newpage
\bibliography{emnlp2018}
\bibliographystyle{acl_natbib_nourl}
\end{document}